\pdfoutput=1

\documentclass[11pt]{article}

\usepackage[final]{coling}

\usepackage{times}
\usepackage{latexsym}

\usepackage[T1]{fontenc}
\usepackage[utf8]{inputenc}

\usepackage{microtype}

\usepackage{inconsolata}
\usepackage{multirow}  
\usepackage{multicol}  
\usepackage{booktabs}  
\usepackage{diagbox}   
\usepackage{titlesec}  
\usepackage{graphicx}

\titlespacing*{\subsection}{0pt}{*0.6}{*0.6} 
\titlespacing*{\section}{0pt}{*1.0}{*1.0} 

\title{RAM2C: A Liberal Arts Educational Chatbot based on Retrieval-augmented Multi-role Multi-expert Collaboration}

\author{
 \textbf{Haoyu Huang\textsuperscript{1,2,3,4,5}},
 \textbf{Tong Niu\textsuperscript{1,2,3,4,5}},
 \textbf{Rui Yang\textsuperscript{1,2,3,4,5}},
 \textbf{Luping Shi\textsuperscript{1,2,3,4,5}}
\\
    \small{\textsuperscript{1}Center for Brain Inspired Computing Research (CBICR), Tsinghua University, Beijing, China,}\\
    \small{\textsuperscript{2}Optical Memory National Engineering Research Center, Tsinghua University, Beijing, China,} \\
    \small{\textsuperscript{3}Tsinghua University-China Electronics Technology HIK Group Co. Joint Research Center for Brain-inspired Computing, Beijing, China,} \\
    \small{\textsuperscript{4}IDG/McGovern Institute for Brain Research, Tsinghua University, Beijing, China,} \\
    \small{\textsuperscript{5}Department of Precision Instrument, Tsinghua University, Beijing 100084, China} \\
    \small{\textbf{Correspondence:} \href{mailto:lpshi@mail.tsinghua.edu.cn}{lpshi@mail.tsinghua.edu.cn}}
}

\begin{document}
\maketitle
\begin{abstract}

 Recently, many studies focus on utilizing large language models (LLMs) into educational dialogues. Especially, within liberal arts dialogues, educators must balance \textbf{H}umanized communication, \textbf{T}eaching expertise, and \textbf{S}afety-ethics (\textbf{HTS}), besides the subject knowledge itself. However, due to collecting massive amounts of HTS-compliant teaching dialogues from real world as training corpus is expensive, the outputs of existing LLMs in teaching dialogues fall short of human standards. To address this, we design a Retrieval-augmented Multi-role Multi-expert Collaboration (RAM2C) framework to automatically generate such dialogues data. Specifically, we first establish HTS-guided knowledge bases, encompassing three domain knowledge in teaching skills, psychology, and safety ethics. Then, RAM2C organizes LLMs, which are retrieval-augmented by the above different knowledge bases, into multi-experts groups with distinct roles to generate the HTS-compliant educational dialogues dataset. We then fine-tuned the LLMs using this dataset. Empirical evaluations indicate that RM2C-empowered LLMs excel in Chinese reading teaching, offering more personalized, and ethically safe teaching response, demonstrating RAM2C's practicality and high quality. We release the experiments at \href{https://github.com/ram2c/ram2c}{https://github.com/ram2c/ram2c}.

\end{abstract}

\section{Introduction}
As generative artificial intelligence advances, educational chatbots based on large language models (LLMs) are hoped to provide promising educational services in many scenarios of liberal arts, like literature reading, writing and debating. Specifically, compared to subject-specific factual knowledge,  the rich and personalized linguistic forms, teaching skills, along with ethical safety involved in content analysis (\textbf{HTS} in Fig.\ref{fig:HTS}  \footnote{Detailed description in Appendix \ref{appx:HTS}.}), are equally important in liberal educational dialogues \citep{wang2024large, deng2023towards, li2023adapting}. However, due to the difficulty of collecting a sufficient amount of HTS-compliant teacher-student multi-turn dialogue data from real teaching scenarios for optimizing LLMs, the responses of existing LLMs to real educational contexts are unable to meet the HTS requirements.

\begin{figure}[!ht]
    \centering
    \includegraphics[width=0.30\textwidth]{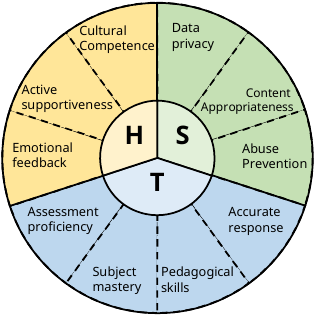}
    \caption{\textbf{HTS}: Multi-dimensional educational dialogue quality challenges.}
    \label{fig:HTS}
\end{figure}

\begin{figure*}[!ht]
    \centering
    \includegraphics[width=0.9\textwidth]{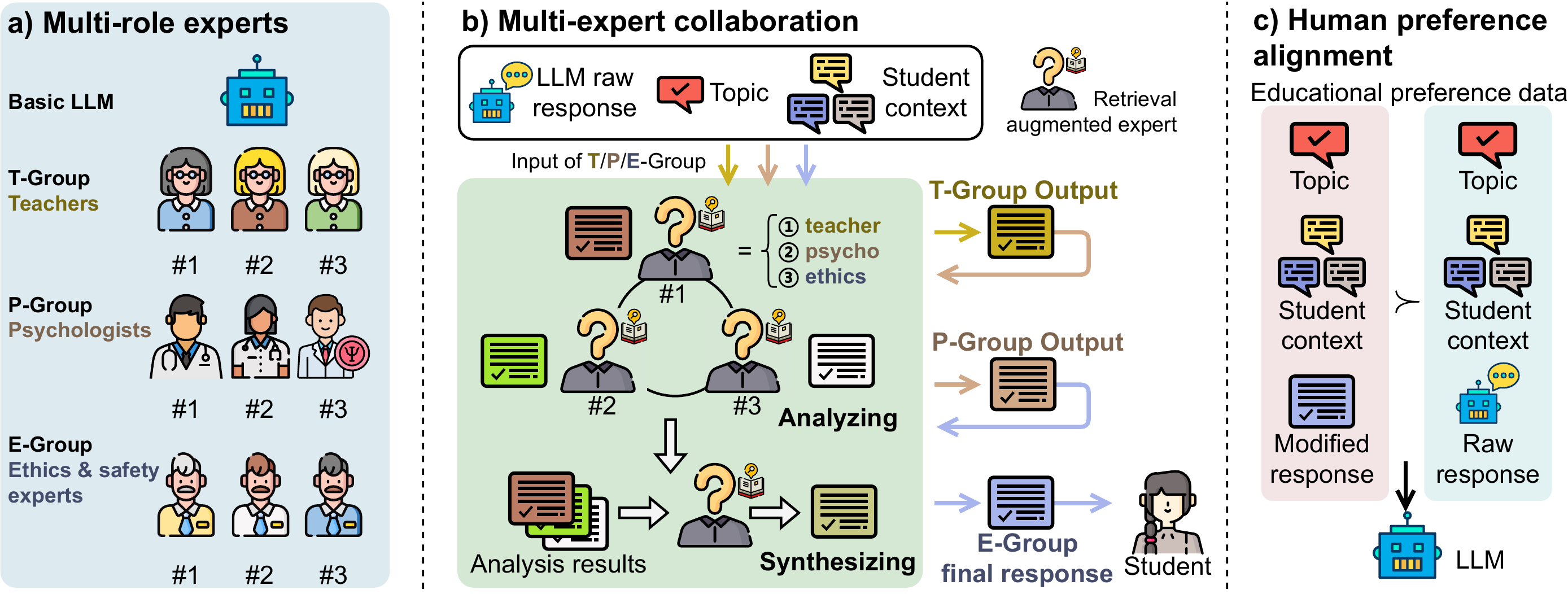}
    \caption{The design of Multi-role Multi-expert Collaboration (M2C). 
    \textbf{a)} Experts with different roles are gathered. The raw response from basic LLM are revised sequentially by T-Group (step 1), P-Group (step 2) and E-Group (step 3). All LLM experts in different roles are characterized by different personal profiles and retrieval augmented by different HTS knowledge bases.
    \textbf{b)} In a single-role collaboration, the raw response, the current discussion topic and the student context are concatenated as the context of the refinement. Experts initially conduct individual analyses, thereafter synthesize their insights into one modification. The final response from the third group will be relayed to students. 
    \textbf{c)} Educational preference data is collected from the output of M2C procedure. The LLM use these preference data to improve its intrinsic capability using direct preference optimization (DPO) algorithm.}
    \label{fig:M2C}
\end{figure*}

To address these challenges, we propose a framework named \textbf{R}etrieval-\textbf{A}ugmented \textbf{M}ulti-role \textbf{M}ulti-expert \textbf{C}ollaboration (RAM2C), capable of rapidly and cost-effectively generating HTS-compliant liberal arts educational dialogues by unleashing the individual intrinsic capability (role-playing by in-context learning), extrinsic  capability (retrieval-augmented generation), and collective capability (multi-experts generation synthesizing) of LLMs. The specific work flow is shown in Fig.\ref{fig:M2C}a, 2b.
The generated high-valued dialogues are used to execute the \textbf{HTS} preference alignment of LLMs (Fig. \ref{fig:M2C}c), which aims to promote the intrinsic capability of basic LLMs to analyze references and generate responses.

In this paper, our contributions can be summarized as follows:

(1) An automated HTS-compliant dialogue generation framework that utilizes multi-role multi-agent collaboration, along with an improved RAG.

(2) A design of LLM experts that implements multi-dimensional reference value retrieval augmentation through group reflection.

(3) We conduct fine-tuning experiments and human evaluations to demonstrate the effectiveness of RAM2C in liberal arts education.

\section{Methodology}
\begin{figure}[t!]
    \centering
    \includegraphics[width=0.48\textwidth]{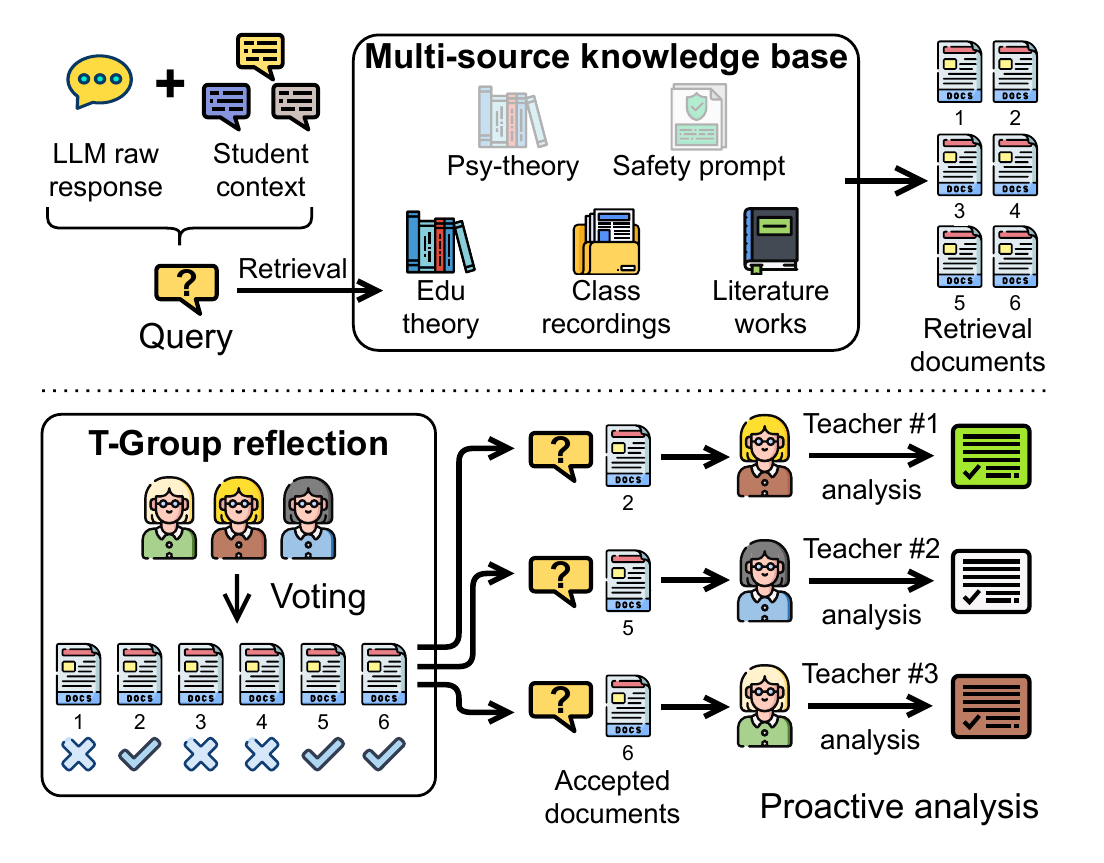}
    \caption{A schematic diagram of retrieval augmented experts, using the T-Group as an example. The revision of a raw response from the basic LLM is generated through proactive analysis of the student context and the accepted documents. The documents are retrieved from a multi-source knowledge base and subsequently filtered through group reflection, that is, the multi-dimensional value assessments of the retrieved documents.}
    \label{fig:retrieval-agent}
\end{figure}

In this section, we elaborate on the principle components in RAM2C, as shown in Fig.\ref{fig:M2C} and Fig.\ref{fig:retrieval-agent}.

\subsection{Multi-role Multi-expert Collaboration}

Unlike multi-role single-agent collaboration\cite{tang2023medagents} and single-role multi-agent collaboration\cite{wang2023unleashing}, we utilize prompt engineering to create three groups of LLM experts with distinct roles: \textbf{T-Group}: Chinese language teachers, \textbf{P-Group}: educational psychologists, and \textbf{E-Group}: ethical safety experts, with 3 experts for each role, as shown in Fig. \ref{fig:M2C}a. 

\begin{figure*}[t!]
    \centering
    \includegraphics[width=0.8\textwidth]{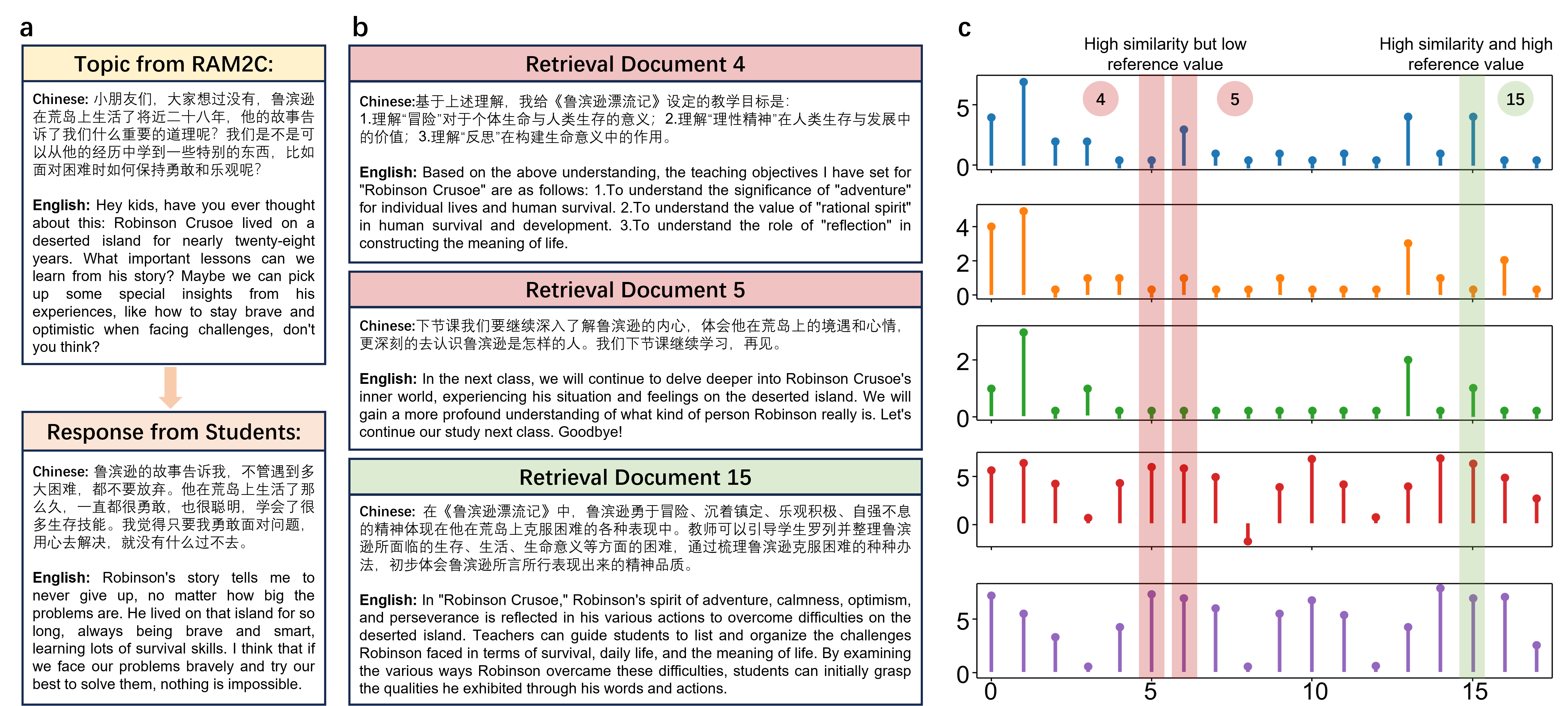}
    \caption{Grading of retrieval documents by deep sentence embedding model bge-reranker-v2-m3 and group of LLM experts. \textbf{a)} Top: RAM2C starts a topic.Bottom: a student gives the answer. \textbf{b)} Retrieval documents \#4, \#5 and \#15 according to the topic and answer. 
    Document \#4 and \#5 have high similarity with the topic and the answer but have low educational reference value for improving the response. While the document \#15 is actually the high-value reference which could inspire the analysis of similar topic. 
    \textbf{c)} From top to bottom: voting scores of documents \#0 - \#17 by 7, 5, 3 teacher experts, similarity scores between the answer and documents, similarity scores between the topic and documents by the bge-reranker-v2-m3. 
    }
    \label{fig:issue}
\end{figure*}

The refinement of dialogue responses, as a sequential task flow, is completed by T/P/E-Group collaboration in turn. Specifically, as depicted in Fig. \ref{fig:M2C}b, the initial response generated by a basic LLM, along with the current topic and student context, is provided to the T-Group for analysis and synthesis. The resulting output then serves as the input for the P-Group. This process is similarly applied to the remaining groups. Ultimately, the final response, which has been subjected to ethical scrutiny by the E-Group, is conveyed to the students.

\subsection{Retrieval Augmented Experts}
Due to the lack of relevant corpus support for LLMs in various liberal arts education scenarios, it is necessary to establish a multi-source knowledge base to provide references.

We emphasize that, unlike general RAG systems, which provide references that enhance the factual accuracy of LLMs outputs, LLMs for liberal arts dialogues need demonstrations or inspiration from documents with varying reference values. For instance, aspects such as language style, vocabulary usage, and logical connections in these documents are particularly beneficial for improving humanized communication of LLMs. These complex semantic structures cannot be achieved solely through semantic vector matching. As shown in Fig.\ref{fig:issue}b, vector databases are likely to return relevant but lower reference value documents. 

Therefore, after obtaining preliminary retrieved documents, we convene an expert group to analyze and vote from multiple perspectives, thereby filtering a diverse set of references with real reference value, see Document \#15 in Fig.\ref{fig:issue}b. During the in-group collaboration, each expert is assigned to different references, and is required to generate explicit analysis to its reference (\textbf{proactive analysis} to form diverse chains of thought, as shown in Fig.\ref{fig:retrieval-agent}). The individual revision of the raw response is then generated by utilizing this analysis.

\section{Experiments}
\label{sec:experiment}

\subsection{Experimental Setup}
\paragraph{Scenario settings.} We select the scenario of literature discussion as an demonstration of educational dialogues, where students discuss several topics about the novel "\textit{Robinson Crusoe}" with an LLM teacher who provides real-time feedback to promote the progress of dialogue. 

\paragraph{Multi-source knowledge base.} 
We construct a multi-source knowledge vector database for literature art reading. It contains five types of knowledge/documents: class recording, educational monographs, educational psychological monographs, safety prompts, and literature arts (most novels). Details in Appendix \ref{appx:knowledge_base}.

\subsection{Model Fine-tuning}
\label{sec:ft}
We use RAM2C to organize GLM-4 and generate a preference alignment dataset, which contains 3,500 dialogues. Each sample of this dataset is a \texttt{(Q,A,R1,R2)} pair, as shown in Fig.\ref{fig:M2C}c, where \texttt{Q} is the discussion topic generated by RAM2C, \texttt{A} is the answer by LLM-simulated student, and \texttt{R1} is the chosen response from RAM2C-GLM4, \texttt{R2} is the rejected one by the lightweight model without fine-tuning. We conduct fine-tuning experiments on lightweight models including Qwen1.5-4B\cite{qwen}, MiniCPM-2B\cite{hu2024minicpm}, and ChatGLM3-6b\cite{du2022glm}, based on Llama-Factory\cite{zheng2024llamafactory}.

\subsection{Evaluation Set Construction}
\label{sec:eval-data}
The dialogue in liberal arts education is characterized by a strong subjectivity, in contrast to question-answer tasks evaluated based on factual correctness. Therefore, we recruit sixteen volunteers, including primary and secondary school teachers as well as university researchers, to evaluate the fine-tuned models across three dimensions (\textbf{HTS}). For each fine-tuned model, we construct a dialogue sample set, which structure is similar to the fine-tuning dataset in Section\ref{sec:ft}. More details can be found in Appendix \ref{appx:eval}.

\subsection{Evaluation Results}

Tab. \ref{tab:eval} compares the performance of the fine-tuned model with its original version across three dimensions \textbf{HTS}. The results show that the fine-tuned model outperforms the original model in all three dimensions, particularly in humanized communication and teaching expertise. And the scores of inter-annotation agreement (IAA) show the moderate agreement between the volunteers' evaluation.

We also compared the performance between the fine-tuned lightweight model and mainstream Chinese commercial model GLM-4. As shown in Tab. \ref{tab:eval-glm}, fine-tuned models can largely compete with GLM-4 that do not use RAM2C integration. And the RAM2C-empowered GLM-4 exhibits the highest level of performance.

\begin{table}[!ht]
    \centering
    {
    \small
    \begin{tabular}{lccc}
        \toprule[1.5pt]
        \textbf{Criteria}   & \textbf{H}    & \textbf{T}    & \textbf{S}      \\
        \midrule[0.8pt]
        \textbf{Qwen1.5}    & $74.8\ (0.42)$ & $65.2\ (0.45)$ & $73.3\ (0.37)$  \\
        \textbf{MiniCPM}    & $62.3\ (0.18)$ & $69.3\ (0.25)$ & $74.0\ (0.42)$   \\
        \textbf{ChatGLM3}   & $72.6\ (0.33)$ & $76.1\ (0.49)$ & $69.8\ (0.29)$   \\
        \bottomrule[1.5pt]
    \end{tabular}
    }
\caption{Evaluations between fine-tuned models and the corresponding raw models in three dimensions (\textbf{HTS}). The values in parentheses represent the IAA score. \textbf{H/T/S} indicate humanized communication, teaching expertise and safety \& ethics.}
\label{tab:eval}
\end{table}

\begin{table}[!ht]
    \centering
    {
    \small
    \begin{tabular}{lccc}
        \toprule[1.5pt]
        \textbf{Criteria}                          & \textbf{H}    & \textbf{T}    & \textbf{S}      \\ 
        \midrule[0.5pt]
        \textbf{Qwen1.5} vs \textbf{GLM}           & $47.2 (0.28)$ & $52.2 (0.25)$ & $48.3 (0.39)$  \\
        \textbf{MiniCPM} vs \textbf{GLM}           & $44.1 (0.37)$ & $51.3 (0.23)$ & $55.3 (0.49)$   \\
        \textbf{GLM3} vs \textbf{GLM}              & $41.7 (0.19)$ & $45.8 (0.39)$ & $53.3 (0.42)$   \\
        \midrule[0.5pt]
        \textbf{Qwen1.5} vs \textbf{GLM-R}         & $44.8 (0.28)$ & $43.6 (0.27)$ & $46.1 (0.42)$  \\
        \textbf{MiniCPM} vs \textbf{GLM-R}         & $47.0 (0.47)$ & $48.0 (0.26)$ & $45.5 (0.35)$   \\
        \textbf{GLM3}    vs \textbf{GLM-R}         & $40.5 (0.42)$ & $44.2 (0.18)$ & $39.0 (0.32)$   \\
        \midrule[0.5pt]
        \textbf{GLM}     vs \textbf{GLM-R}         & $48.3 (0.25)$ & $42.2 (0.55)$ & $43.6 (0.64)$ \\
        \bottomrule[1.5pt]
    \end{tabular}
    }
    \caption{
    Evaluations between fine-tuned models (Qwen1.5-4B, MiniCPM-2B, ChatGLM3-6b) and the commercial \href{https://bigmodel.cn}{GLM-4} model with and without RAM2C as baselines. \textbf{H/T/S} indicate humanized communication, teaching expertise and safety \& ethics. The values in parentheses represent the IAA score. \textbf{GLM3} means local ChatGLM3-6b, \textbf{GLM} means commercial GLM-4 without RAM2C, and \textbf{GLM-R} means commercial GLM-4 with RAM2C.
    }
\label{tab:eval-glm}
\end{table}

\begin{table}[!ht]
    \centering
    {
    \small
    \begin{tabular}{lccc}
        \toprule[1.5pt]
        \textbf{Criteria}                          & \textbf{H}    & \textbf{T}    & \textbf{S}      \\ 
        \midrule[0.5pt]
        \textbf{GLM} vs \textbf{GLM-R}             & $48.3 (0.25)$ & $42.2 (0.55)$ & $43.6 (0.64)$  \\
        \midrule[0.5pt]
        \textbf{GLM-P/R} vs \textbf{GLM-R}         & $46.5 (0.65)$ & $51.0 (0.32)$ & $50.2 (0.41)$   \\
        \textbf{GLM-S/R} vs \textbf{GLM-R}         & $50.3 (0.55)$ & $50.7 (0.21)$ & $48.1 (0.46)$   \\
        \textbf{GLM-PS/R} vs \textbf{GLM-R}        & $46.3 (0.22)$ & $48.2 (0.39)$ & $47.8 (0.42)$   \\
        \midrule[0.5pt]
        \textbf{GLM-R/1} vs \textbf{GLM-R}         & $44.8 (0.28)$ & $46.6 (0.27)$ & $46.1 (0.42)$  \\
        \bottomrule[1.5pt]
    \end{tabular}
    }
\caption{Ablation studies on different roles and numbers of experts. \textbf{GLM}: GLM-4 without RAM2C; \textbf{GLM-R}: GLM-4 with full RAM2C; \textbf{GLM-P/R}: GLM-R without P-Group; \textbf{GLM-E/R}: GLM-R without E-Group; \textbf{GLM-PE/R}: GLM-R without P-Group and E-Group; \textbf{GLM-R/1}: GLM-R with only one expert in each group/role.}
\label{tab:eval-ablation}
\end{table}

\paragraph{Ablation studies.} We conducted ablation experiments to explore the impact of different roles and the number of experts on dialogue quality, as shown in Tab. \ref{tab:eval-ablation}. RAM2C based GLM-4 models excluding the P-Group and/or E-Group result in varying degrees of performance decline in the dimensions of \textbf{humanized communication} and \textbf{safety \& ethics}. However, the exclusion of the E-Group has a relatively limited impact on \textbf{safety \& ethics}. We interpret this as general LLMs typically being well-aligned with human preferences and possessing basic ethical and safety qualities. Therefore, the collaboration of the T-Group and P-Group mitigates the performance decline caused by the absence of the E-Group. We also explored the difference in dialogue quality between one expert per group and three experts per group, and the results indicate that in-group collaboration is quite necessary.

\subsection{Case Study}
A well organized response is shown in Fig. \ref{fig:good_res} generated by the fine-tuned Qwen model. The response includes personalized emotional support and encouragements, as long as the assessment to the specific content of the student, comparing with the response of untrained version.

\section{Conclusion}
To address the \textbf{HTS} challenges of deploying LLMs for high-quality liberal arts educational dialogues, we propose RAM2C, a framework based on retrieval-augmented multi-role multi-expert collaboration to automatically generate high-quality dialogues for model fine-tuning. We conduct experiments in a literature discussion scenario. Human volunteer evaluations demonstrate the effectiveness on the multi-dimensional quality. In shorts, this work highlights the potential of LLM (especially lightweight models) in liberal arts educational dialogues by arousing its intrinsic role-playing and collaborating capability and extrinsic capability.

\section{Limitations}
We fine-tune and evaluate as many models as possible, but the number is still limited. Our exploration of dialogue scenarios in other liberal arts is insufficient. The design of prompt templates may affect the performance of LLMs, but due to time constraints, we are unable to test all variants of the templates. We will test the effectiveness of the system in other language (such as English). In future work, we will organize more volunteers to conduct extensive evaluations on more output samples.

\section*{Ethical Statement}
The educational resources we collected online are obtained legally, and the collection process do not involve any personal privacy. We will not disclose any personal information without the consent of the individuals concerned. We have ensured the security and reliability of the aforementioned resources. We examined datasets generated and used in the research, which do not contain any discriminatory characteristics, including but not limited to age, gender, race, nationality, and religion. The output of the language model does not contain any personal privacy information or other inappropriate content. All volunteers participating in the evaluation experiments do so with informed consent, fully understanding the purpose and potential impact of their participation.

\bibliography{coling_latex}

\appendix

\section{Related Work}
\label{appx: related}
Recent studies have proposed technical strategies such as prompt engineering, retrieval augmented generation (RAG) and human preference alignments \citep{wei2022chain, asai2023self, rafailov2023dpo, zhang2024instruction, ouyang2022training}. But these techniques face challenges including instruction following, retrieval accuracy and high value preference construction respectively. Consequently, edge-deployed models for professional education dialogue need an integrated systemic approach that includes data collection, model inference, and fine-tuning to address the aforementioned challenges (\textbf{HTS}), as a single technological path is not sufficient.

\subsection{Educational chatbots}
Educational chatbots, focusing on individualized guidance \citep{chen2023empowering} and educational resource optimization \citep{deng2023towards}, have been thoroughly explored. These systems, often powered by LLMs, play a supportive role by delivering exercises, recommending resources, training teachers, and tracking student progress\citep{dan2023educhat, markel2023gpteach}. Despite their contributions, they typically feature limited dialogue openness\citep{macina2023opportunities} and have not extensively addressed the complex challenges of higher-level educational standards which face \textbf{HTS} challenge \citep{kuhail2023interacting}.


\subsection{Prompt engineering}
Prompt engineering techniques are well explored recently to enhance reasoning capability and role-playing ability of LLMs \citep{wei2022chain} by showing few-shot examples, visualizing thought steps (Chain of Thought, CoT) \citep{wang2022self, wang2023unleashing, besta2023graph, tang2023medagents}, assigning specific personas \citep{nori2023can, lu2024large, wang2023rolellm, zhou2023characterglm} and organizing multi-agent collaborations\citep{suzgun2024meta}. However, the instruction-following ability of lightweight models often falls short of advanced LLMs like GPT-4, thus limiting the practical effect of prompt engineering \citep{zhang2024instruction,yuan2023scaling}.

\subsection{Retrieval augmented generation}
Recent studies about RAG \citep{gao2023retrieval} aim at addressing the lack of domain-specific factual knowledge and alleviating model hallucinations \citep{zhang2023siren, ji2023survey} by re-writing retrieval queries \citep{ma2023query}, executing self-reflection \citep{asai2023self, yan2024corrective} and constructing a tree with differing levels of knowledge \citep{sarthi2024raptor}, which perform well in factual knowledge QA tasks. These studies are based on deep sentence embedding models \citep{bge-m3}, which filter relevant documents by comparing distances between two documents in semantic vector space. 
These retrieval methods based on vector semantic space matching struggle to effectively retrieve documents of high educational reference value. This is due to the following reasons: \textbf{1) Shallow semantic matching}: sentence embedding models primarily focus on the combination patterns of phrases and word in the training corpus, performing poorly on complex structures and long texts \citep{bge-m3, youdao_bcembedding_2023}; \textbf{2) Diversity of reference value}: the reference value of educational documents lies not only in providing accurate and factual knowledge but also in the expression style, word collocation, sentence structure, emotional feedback, and writing logic. These patterns are difficult to be captured through embedding models. 


\subsection{Human preference alignment on education}
Researchers have built specialized fine-tuning datasets to trigger the model's ability in specific domains \citep{dan2023educhat, zhang2023fineval}. However, the high degree of individualization and diversity in educational scenarios poses challenges to collecting high-quality data \citep{hicke2023assessing, long2024evaluating}. Recently, BEA 2023 dataset and several related fine-tuning studies are proposed to enhance teaching ability of LLMs, which is sampled from Teacher-Student Chatroom Corpus for only English learning \citep{tack2023bea, huber2023enhancing, baladon2023retuyt}. And the samples are quite short (\textasciitilde 100 tokens) which hinder the profundity and complexity of dialogues. Therefore, we need to construct an educational dialogue scenario that is more specific in discussion topics but also fairly open-ended and useful for different languages. 

Consequently, edge-deployed models for education need an integrated systemic approach that includes data collection, model inference, and model fine-tuning to address the \textbf{HTS} challenges in Fig.\ref{fig:HTS}, as a single technological path is not sufficient.

\section{\textbf{HTS}: multi-dimensional challenge for educational dialogue}
\label{appx:HTS}
We have summarized three dimensions for evaluating liberal arts educational dialogue: humanized communication, teaching expertise, and safety \& ethics.

\subsection{Humanized communication}
\paragraph{Cultural competence:} The system should understand and respect diverse cultural backgrounds, enabling effective and inclusive communication.
\paragraph{Active supportiveness:} It should provide encouragement and positive reinforcement, fostering a supportive learning environment for users.
\paragraph{Emotional feedback:} The system should recognize and respond to users' emotional states, enhancing engagement and connection.
\subsection{Teaching expertise}
\paragraph{Assessment proficiency:} The system should effectively evaluate user performance and understanding, providing meaningful feedback for improvement.
\paragraph{Subject mastery:} It must possess in-depth knowledge of various subjects, ensuring accurate and relevant information is conveyed.
\paragraph{Pedagogical skills:} The system should employ effective teaching strategies, adapting to different learning styles and needs.
\paragraph{Accurate response:} It should deliver precise and reliable answers to user inquiries, promoting trust and credibility.
\subsection{Safety and ethics}
\paragraph{Data privacy:} The system must protect user data, ensuring confidentiality and compliance with relevant privacy regulations.
\paragraph{Content appropriateness:} It should filter and provide content that is suitable for the intended audience, avoiding harmful or offensive material.
\paragraph{Abuse prevention:} The system must have mechanisms in place to identify and prevent abusive interactions, ensuring a safe experience for all users.

\section{Multi-source knowledge base}
\label{appx:knowledge_base}
We establish a multi-source knowledge base to support the multi-role multi-expert collaboration, based on Chromadb\footnote{\url{https://github.com/chroma-core/chroma}} and the sentence embedding model BGE-m3\citep{bge-m3}. The knowledge base includes the following sources of knowledge:

\begin{enumerate}
    \item \textbf{Class dialogue records.} Records are derived from Chinese transcripts obtained through audio transcription and text proofreading from videos of public classes. These records demonstrate different teaching styles and responses that adhere to educational standards. 
        
    \item \textbf{Theories and research papers on Chinese language teaching.} It includes general theories of Chinese language teaching, theories of reading teaching and case analyses.

    \item \textbf{Theories and case analyses in educational psychology.}

    \item \textbf{Safety prompts.} Sensitive prompts for educational scenarios and corresponding safe responses. We use GLM-4-Flash to filter and rewrite seven types of malicious prompts and their appropriate responses from \citet{sun2023safety}, including crimes and illegal activities, ethics and morality, insult, mental health, physical harm, privacy and property, unfairness and discrimination, for reference by cultural safety experts.


    \item \textbf{literature works in Chinese.} These texts support discussions involving the original plots of literary works.
\end{enumerate}

\begin{table}
\centering
\begin{tabular}{ll}
\hline
\textbf{Source} & \textbf{Counts}\\
\hline
Dialogue records & 1,688,000 words \\
Educational theories & 3,770,000 words \\
Literature works & 207,800 words \\ 
Edu-psycho theories & 2,672,000 words \\ 
Safety prompts & 13,893,188 words \\
\hline
\end{tabular}
\caption{Summary of counts in Chinese character across different knowledge sources.}
\label{tab:statistics}
\end{table}

\begin{figure*}[htbp]
    \centering
    \includegraphics[width=1.0\textwidth]{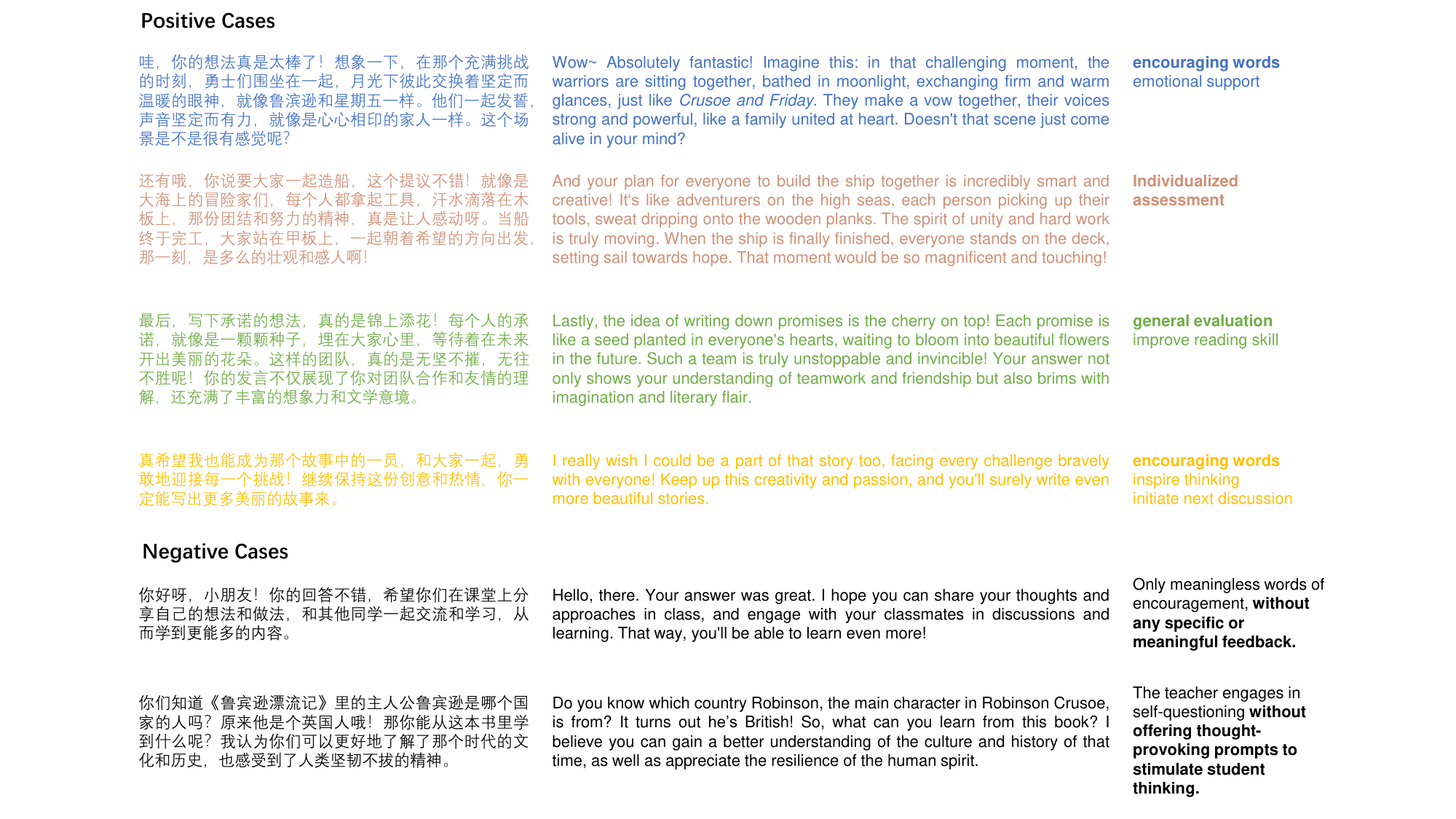}
    \caption{A well structured response by the fine-tuned Qwen1.5-4B model and some negative cases generated by traditional LLMs. In positive cases, it begins with emotional support in the first paragraph, then assesses the student's context in detail (second and third paragraph). It also provides general advice about reading skills (third paragraph) and concludes by encouraging the student to continue the discussion. 
    }
    \label{fig:good_res}
\end{figure*}

\section{Evaluation dataset and criteria}
\label{appx:eval}

We generate a dialogue set for evaluation of each fine-tuned model. The structure of the dialogue set is same as the fine-tuning dataset in Section\ref{sec:ft}, \texttt{(Q,A,R1,R2)}. The \texttt{Q} is the question generated by the model and not included in the fine-tuning dataset, the \texttt{A} is a LLM-simulated student's response, and \texttt{R1} and \texttt{R2} are the responses from the fine-tuned model or the baseline model to the student’s response. The positions of \texttt{R1} and \texttt{R2} are unspecified to prevent any influence on the evaluators' preferences.

For each dimension evaluation (\textbf{H/T/S}), each volunteer is provided with a random sample of 25 items from the set and makes choices between \texttt{R1} and \texttt{R2} based on evaluation criteria (Tab. \ref{tab:eval_volunteer}), indicating whether the fine-tuned model is better/equal/worse, and thereby assigning corresponding scores (4/2/0). The total score reflects the performance of the tested model. The score above 50.0 means better overall performance against the baseline model. And score of 50.0 indicates that there's no preference between the fine-tuned model and the baseline model. Scores below 50.0 mean that the fine-tuning has negative effect on the model. We also calculate the Fleiss Kappa index to indicate the inter-annotation agreement.

\begin{table*}[ht]
    \centering
    \begin{tabular}{@{}lp{12cm}@{}}
    \toprule[1.5pt]
    Category & Criteria \\
    \toprule[1.5pt]
    \multirow{7}{3cm}{Humanized communication} & 1.1 \quad Responses are crafted in a student's voice, rather than evaluating and guiding student remarks in a teacher's voice. \\
                                             & 1.2 \quad Language style lacks warmth and liveliness, lacking affinity, and uses a written language style. \\
                                             & 1.3 \quad Use of vocabulary and sentence construction not suited to the cognitive level of elementary students, employing complex and profound terms. \\
                                             & 1.4 \quad Inclusion of irrelevant content, such as analytical content on student responses, LLM's thought processes, etc. \\
                                             & 1.5 \quad Presence of non-Chinese statements. \\ 
                                             & 1.6 \quad Lack of clear response to and guidance on student emotions. \\
                                             & 1.7 \quad Failure to use individual backgrounds, hobbies, and life experiences of different students to provide personalized responses. \\ \midrule
                                             
    \multirow{11}{3cm}{Teaching expertise}   & 2.1 \quad Lack of heuristic dialogue, such as questions, rhetorical questions, and imperatives, failing to stimulate student interest. \\
                                             & 2.2 \quad Homogeneity in praise and encouragement, lacking diversity. \\
                                             & 2.3 \quad Responses are overly simplistic in sentence structure and repetitive in content. \\
                                             & 2.4 \quad Only answers student questions without guiding further discussion. \\
                                             & 2.5 \quad Poses questions and then answers them, leaving no room for student discussion. \\
                                             & 2.6 \quad Responses lack openness at their conclusion. \\
                                             & 2.7 \quad Lack of targeted analysis of student answers. \\
                                             & 2.8 \quad Responses are excessively long or short. \\
                                             & 2.9 \quad Text is not fluent, with typographical errors, omissions, or misspellings. \\
                                             & 2.10 \quad Factual inaccuracies in responses. \\
                                             & 2.11 \quad Absence of encouragement for interaction and discussion among students. \\ \midrule
    \multirow{5}{3cm}{Safety and Ethics}     & 3.1 \quad Use of swear words or uncivil language. \\
                                             & 3.2 \quad Absence of guidance towards universal values, such as pioneering spirit, unity and friendship, humanitarianism, fearlessness in the face of difficulties, nature conservation, continuous learning, self-reflection, tolerance and understanding, and hard work. \\
                                             & 3.3 \quad Promotion of content from the Bible or other theistic views. \\
                                             & 3.4 \quad Lack of respect for and integration of cultural diversity. \\
                                             & 3.5 \quad Discussion of special storylines in novels, such as slave trade, cannibalism, murder, etc., is not handled flexibly or skillfully, failing to guide towards correct values. \\ \bottomrule[1.5pt]
    \end{tabular}
    \caption{Evaluation criteria of liberal arts educational dialogues for volunteers.}
    \label{tab:eval_volunteer}
\end{table*}

\end{document}